# Toward a Period-Specific Optimized Neural Network for OCR Error Correction of Historical Hebrew Texts

OMRI, O.S, SUISSA

Bar Ilan University, Department of Information Science, Ramat Gan 52900, Israel

MAAYAN, M.ZG, ZHITOMIRSKY-GEFFET

DR, Bar Ilan University, Department of Information Science, Ramat Gan 52900, Israel

AVSHALOM, A.E, ELMALECH

DR, Bar Ilan University, Department of Information Science, Ramat Gan 52900, Israel

Over the past few decades, large archives of paper-based historical documents, such as books and newspapers, have been digitized using the Optical Character Recognition (OCR) technology. Unfortunately, this broadly used technology is error-prone, especially when an OCRed document was written hundreds of years ago. Neural networks have shown great success in solving various text processing tasks, including OCR post-correction. The main disadvantage of using neural networks for historical corpora is the lack of sufficiently large training datasets they require to learn from, especially for morphologically-rich languages like Hebrew. Moreover, it is not clear what are the optimal structure and values of hyperparameters (predefined parameters) of neural networks for OCR error correction in Hebrew due to its unique features. Furthermore, languages change across genres and periods. These changes may affect the accuracy of OCR post-correction neural network models. To overcome these challenges, we developed a new multi-phase method for generating artificial training datasets with OCR errors and hyperparameters' optimization for building an effective neural network for OCR post-correction in Hebrew. To evaluate the proposed approach, a series of experiments using several literary Hebrew corpora from various periods and genres were conducted. The obtained results demonstrate that (1) training a network on texts from a similar period dramatically improves the network's ability to fix OCR errors, (2) using the proposed error injection algorithm, based on character-level period-specific errors, minimizes the need for manually corrected data and improves the network accuracy by 9%, (3) the optimized network design improves the accuracy by 3% compared to the state-of-the-art network, and (4) the constructed optimized network outperforms neural machine translation models and industry-leading spellcheckers. The proposed methodology may have practical implications for digital humanities projects that aim to search and analyze OCRed documents in Hebrew and potentially other morphologically-rich languages.

**CCS CONCEPTS** • Computer systems organization~Architectures~Other architectures~Neural networks • Applied computing~Document management and text processing~Document capture~Optical character recognition • Computing methodologies~Artificial intelligence~Natural language processing~Natural language generation

# 1 INTRODUCTION

In the last few decades, large archives of paper-based historical documents, such as books and newspapers, have been digitized using the OCR (Optical Character Recognition) technology. For example, the Google Books project [Heyman, 2015] and the British Newspaper Archive (https://www.britishnewspaperarchive.co.uk/) maintain extensive digitized collections with advanced discovery tools [Lansdall-Welfare, Sudhahar, Thompson, Lewis, & Cristianini, 2017]. OCR algorithms convert a high-resolution image of the printed textual resource into machine-encoded text. The quality of the obtained OCRed text is critical for both research and preservation of the historical and cultural heritage conveyed by these resources. Unfortunately, while commercial OCR products achieve good results, they are not 100% accurate, especially when applied to historical texts. OCR errors, sometimes called spelling mistakes, come in several forms: insertions, deletions, substitutions, transposition of characters, and spacing errors that lead to splitting and combining of words [Reynaert, 2008; Jatowt, Coustaty, Nguyen & Doucet, 2019].

In order to automatically correct OCR errors in large historical texts, supervised machine learning algorithms have been applied [Raaijmakers, 2013; Evershed & Fitch, 2014; Kissos & Dershowitz, 2016]. One of the most effective machine learning approaches that have shown great success in solving various text processing tasks, including spell checking [Raaijmakers, 2013] and machine translation [Mokhtar, Bukhari & Dengel, 2018] are deep neural networks (DNN). Recently, several initial DNN-based techniques for post-correction of OCR errors have been devised [Chiron, Doucet, & Moreux, 2017; Rigaud, Doucet, Coustaty & Moreux, 2019]. There are many different hyperparameter configurations and network structures that can be used to construct an effective DNN for this task. Systematically checking all possible combinations of different values of hyperparameters would result in tens of thousands of network structures. It would take years to compute all of them, in order to find the optimal neural network design. Therefore, there is a need for a more practical and efficient approach for network optimization. Moreover, building DNN for correcting OCR errors requires a vast amount of training data (i.e., OCRed corpora with parallel gold standard textual versions). There is a lack of such training datasets for historical texts in various national languages, and, specifically in Hebrew. In addition, to the best of our knowledge, there is no previous research on the utilization of DNN for fixing OCR errors in Hebrew.

Hebrew is a morphologically-rich language with unique characteristics that make it hard to analyze using machine learning algorithms [Tsarfaty, Seddah, Goldberg, Kübler, Versley, Candito, & Tounsi, 2010]. Unlike simple morphology languages, such as English, morphologically-rich languages are more complex to analyze using standard NLP pipelines due to high ambiguity level [Tsarfaty, Seker, Sadde, & Klein, 2019]. For example, using standard pipelines, such as spaCy, Stanford's CoreNLP [Manning, Surdeanu, Bauer, Finkel, Bethard, & McClosky, 2014] or a WordPiece technique (common DNNs using transformers) [Wu, Schuster, Chen, Le, Norouzi, Macherey, & Dean, 2016] for the Hebrew word City (עיר) will produce about 500 "morphologically similar" words. These words include Cities (ערים), City (עיר), Young (צעיר), Chives (עירית), Alertness (עירנות), Naked (עירום), Iraq (עיראק) and others. Most of these words are not correct derivations of the word עיר and are not semantically related to its meaning, consequently reducing the accuracy of the OCR post-correction. But all of them have the same (or similar) root word under simplistic assumptions (i.e., English and other Latin-based languages NLP assumptions), but not according to the Hebrew language morphological rules. A root word is a fundamental element of Semitic languages, such as Hebrew and Arabic, that allows attaching prefixes or suffixes to a lexical unit (word or sub-word) to create different forms and grammatical derivations while preserving a similar semantic meaning. In addition, Hebrew prepositions (e.g., in, on, at) are prefixes rather than separate words adding more complexity to automatic text processing and word sense disambiguation. Moreover, since Hebrew is in use for thousands of years, it continuously changes and evolves across periods and genres [Hoffman, 2004]. The historical Hebrew texts were also affected by the wide geographical spread of the Jewish people during the last two millennia. Therefore, it is unclear whether state-of-the-art NLP methods for various tasks, and specifically for OCR post-correction that work well on other languages [e.g., Hämäläinen, & Hengchen, 2019] will produce good results for historical Hebrew texts.

This research aims to investigate the effect of rich morphology as well as the influence of period and genre on the accuracy of neural network models for OCR post-correction in Hebrew and propose a generic and efficient approach to design an effective model for this task. Although the proposed methodology was tested on historical newspapers in Hebrew, we believe that it can be generalized and applied to historical corpora from other genres and periods in different morphologically-rich languages.

The main research questions addressed in this research are:

1. How to efficiently design an optimized DNN for error correction of OCRed texts with minimal computing and human effort?
2. How does the type of training dataset (period and genre) influence the accuracy of the network's corrections in morphologically-rich languages?

To this end, a greedy approach for optimizing DNN's hyperparameters was used based on submodular optimization [Jin, Yan, Fu, Jiang & Zhang, 2016]. The greedy approach is computationally efficient since it tests one hyperparameter at a time and fixes its optimal value for the next hyperparameter tests. To minimize the amount of manually created training data, an OCR error generation algorithm was developed using common errors that are typical for a given historical Hebrew corpus based on the analysis of a few manually annotated articles. This algorithm automatically generated more training data by inserting common OCR errors into correct historical Hebrew texts.

To test the proposed methodology, the network constructed by the greedy optimization approach was trained on different historical Hebrew corpora with artificially generated OCR errors. The quality of the obtained results was evaluated on a subset of OCRed Hebrew historical newspapers from JPress (http://web.nli.org.il/sites/JPress/Hebrew/Pages/default.aspx). To show that the proposed methodology might be successfully applied to other morphologically-rich languages, we also compared the performance of the suggested optimized model with one of the best models from ICDAR 2019 [Rigaud et al., 2019] on the historical Polish training dataset.

## 2 RELATED WORK

### 2.1 OCR Post-Correction

OCR post-correction has recently become a critical task in many digitization projects. Several types of OCR errors have been examined in the literature [Fernandes, Santos, & Milidiú, 2010]. The occurrence and frequency of these errors depend on the period, genre, language, and other features of the text. Popular approaches for the correction of OCR errors that have been explored in previous research were based on crowdsourcing [Suissa, Elmalech, & Zhitomirsky-Geffet, 2019], "noisy channel" [Kolak & Resnik, 2002], language models [Evershed & Fitch, 2014], and machine learning [Kissos & Dershowitz, 2016] (including deep learning [Rigaud et al., 2019]) as described next.

One of the most trivial methods for correcting OCR errors is the use of crowd-workers. Crowdsourcing is based on assigning various well-defined tasks to large groups of non-expert workers or volunteers. Using crowdsourcing platforms, workers worldwide can be recruited to check and fix OCR errors efficiently and effectively. For instance, [Volk, Clematide, & Furrer 2016] presented a scanned image of a document along with the OCRed text and asked the crowd workers to correct errors [Volk et al., 2016]. To keep the task as simple as possible, some studies split the proofing task into sub-tasks of error finding, fixing, and verifying corrections [Bernstein et al., 2010]. This approach outperforms automatic spellchecking algorithms and significantly improves the quality of the resulting text. An optimal crowdsourcing methodology for this task was proposed in a recent study [by Suissa et al., 2019]. Although this method yields improved results, it mostly fits widely spread languages like English and Spanish. Hebrew and many other national languages do not have enough speakers for using such an approach effectively. For example, in Amazon Mechanical Turk (the most popular crowdsourcing platform globally), only the following national languages are available (English is assumed as mandatory): Portuguese, Chinese/Mandarin, French, German and Spanish. Moreover, when using a paid platform like Amazon Mechanical Turk, OCR large post-correction projects can be quite costly, while on a voluntary platform, they tend to progress very slowly. Therefore, in practice, the applicability of the crowdsourcing-based approach for OCR error post-correction is limited, especially for morphologically-rich languages like Hebrew.

An alternative approach for OCR post-correction can be the noisy channel method. Noisy channel models treat the OCR error token (i.e., word) as if a correctly spelled word had been "distorted" by being passed through a noisy communication channel. The "noise" is the changes that have been made to the correction in the "communication channel". This noisy channel model is a kind of Bayesian inference; given an OCR error, the model needs to find a corresponding correct word out of a given vocabulary (i.e., the word with the highest probability to be the corrected word) [Jurafsky & Martin, 2016]. The most significant disadvantage of the noisy channel method is the fact that it cannot handle "out of vocabulary" words. Morphologically rich languages are more vulnerable to "out of vocabulary" scenarios due to a large fraction of letter combinations that form valid words.

Traditional machine learning approaches for OCR post-correction show some success on morphologically rich languages [Kissos, & Dershowitz, 2016]. However, the traditional machine learning methodology requires manual selection and modeling of features that may represent OCR post-correction errors. Common features that were found to be useful include confusion weight (the number of occurrences of a word in a corruption-correction confusion matrix), "noisy channel" probabilities, unigram frequency of the word in the corpus, bigram frequency of the correction and other correction candidates in the corpus. Unfortunately, it is not easy to manually identify all the features that may represent OCR errors in terms of textual elements. Moreover, even when such features are found and proven to be effective, there are still may be others that have been missed by the researcher. Deep learning solves this central problem by automatically learning feature representations based on examples instead of using explicitly predefined features [Deng & Liu, 2018].

## 2.2 OCR Post-Correction using Deep Learning

Deep learning is a machine learning approach that uses deep neural networks (DNN) models. DNN is a computational mathematical model that consists of several "neurons" arranged in layers. The network receives an input in the first layer of neurons, and the input is passed over and changed from layer to layer until it becomes the output in the last layer of the network. Changes in the weights of the network are usually performed using the Gradient Descent method [Barzilai & Borwein, 1988]. The learning process is divided into two main parts: epoch and batch. The number of samples the network uses before calculating the gradients is the batch size. The epoch consists of several batches, and its size is determined by the size of the training dataset. The output of each neuron, the neurons to which it transmits the output, and additional parameters make up a neuron layer. Many hyperparameters of the network, such as the number of layers, number of neurons in a given layer, batch size, learning rate, and others, cannot be learned by the network and have to be predefined and tuned. To tune or optimize the hyperparameters, the network should be repeatedly trained on some initial hyperparameters' values and evaluated on a validation set. The values are modified after each cycle, aiming to achieve the best possible results with the shortest learning time.

Over the years, different deep learning layers have been developed with various abilities. A deep learning layer, such as LSTM (Long Short-Term Memory) [Hochreiter & Schmidhuber, 1997], allows the network to "remember" previously calculated data (a character in this study) and thus learn corrections with respect to an entire sequence (a sentence in this study). Another commonly used layer is Gated Recurrent Units (GRU) [Cho et al., 2014]. Both the LSTM and the GRU layers can be implemented in the direct and reverse order of the sequence. This technique is called bidirectional-RNN [Schuster & Paliwal, 1997] and allows the layer to learn a broader context by learning to correct by the following words rather than only by the preceding words. This technique transfers the data twice within the network: in the direct and the opposite order of the sequence (from the end to the beginning of the sequence).

These layers are used to construct different models, including a sequence-to-sequence model [Sutskever, Vinyals, & Le, 2014] that uses an encoder-decoder [Cho et al., 2014] architecture. In addition, the number of layers required for the encoder and decoder depends on the type and complexity of the task. To prevent overfitting and overcome the vanishing gradient problem [Hochreiter, 1998], a "dropout" regulation layer is added [Srivastava, Hinton, Krizhevsky, Sutskever, & Salakhutdinov, 2014]. In each learning process, several neurons are discarded so that the network must learn to generalize and cannot use a simple mapping between the input and the output.

In recent years, deep learning models applying various DNN have been found very effective in many natural language processing tasks. In particular, the correction of spelling errors [Raaijmakers, 2013], translation [Mokhtar, et al., 2018], and learning editing operations [Chrupała, 2014] are performed using these models with great accuracy. Therefore, using deep learning to fix OCR errors seems like a promising direction, especially the encoder-decoder architecture that is a critical part of the Neural Machine Translation (NMT) approach, which maps (i.e., translates) a sequence input (e.g., a sequence of OCRed characters with errors) to a sequence output. The encoder encodes the sequence into a fixed-size context vector used by the decoder to predict the output (e.g., the correct sequence of the characters). A recent competition in the field illustrates the wide range of possible approaches [Chiron, Doucet, & Moreux, 2017; Rigaud, et al., 2019], where the common trend seems to be the integration of neural networks in most solutions.

Since neural networks learn only from examples in the training corpus, and there is no way to predefine any specific rules, there is a crucial impact of the language and variety of texts in the training corpus. DNN shows great success in learning language models using language-specific training datasets [Mikolov, Deoras, Povey, Burget, & Černocký, 2011]. Some NLP tasks can be solved using language-agnostic approaches [Wehrmann, Becker, Cagnini, & Barros, 2017]. However, language-agnostic models assume that different languages have the same level of complexity (need the same number of neurons to model the language features), which in many cases, is not valid [Tsarfaty, Seker, Sadde, & Klein, 2019]. Therefore, it seems critical to adjust the training data and even the network structure to the targeted language to provide optimal results. Various machine learning and deep learning approaches for OCR post-correction of historical texts have been investigated in different languages, such as Finnish [Drobac, Kauppinen, & Linden, 2017; Hakala, Vesanto, Miekka, Salakoski, & Ginter, 2019], Vietnamese [Nguyen, Le, & Zelinka, 2019], Bengali [Hasnat, Habib, & Khan, 2008] and others [Amrhein & Clematide, 2018; Mokhtar et al., 2018; Krishna, Majumder, Bhat, & Goyal, 2018; Nguyen, Jatowt, Nguyen, Coustaty, & Doucet, 2020]. These studies used a DNN classifier to classify OCR errors, dictionary-based OCR error detection, modeling linguistic features of OCR errors, Hidden Markov Model (HMM) for character classification, statistical machine translation (SMT), natural machine translation (NMT), and transformers approaches. For example, in ICDAR 2019, a team from the Indian Institute of Technology in Bombay used OpenNMT [Klein, Kim, Deng, Senellart, & Rush, 2017] (https://opennmt.net/) in a method they called Character Level Attention Model (CLAM). Although NMT at a character level seems to be the leading approach, there are still many different variations of DNN models [Rigaud et al., 2019], and there is no optimal methodology for building an optimized DNN for this task. Moreover, as discussed above, morphologically-rich languages have unique features that may impact the OCR post-correction process.

## 3 METHODOLOGY

### 3.1 Optimized Training Dataset Generation for DNN

To build an optimal DNN model for a specified task, a network has to be trained on a large dataset for which both the input (OCRed text with errors) and the target data (manually fixed golden standard text) are provided. Such datasets are often unavailable for historical texts in small-scale national languages like Hebrew. Thus, to minimize the amount of manually fixed OCRed texts required for training DNN, we developed a method for training data generation by inserting artificially generated OCR errors into correct texts. We experimented with two types of errors: 1) generic (period-agnostic) errors; 2) period-specific errors.

Generating a period-specific training dataset requires some knowledge about the common OCR errors typical for the language of the corpus. The proposed method for generating and using an artificial training dataset consists of the following main stages:

(1) Collecting a minimal number of texts with OCR errors and their corresponding fixed texts (golden standard) from the examined language, genre, and period. To fix the texts and create a golden standard, experts' work or crowdsourcing can be exploited as presented in the previous work [Suissa et al., 2019];

(2) Sampling errors in the above texts by comparing them to the corresponding fixed versions and mapping error types and their frequencies in these texts [Kumar and Lehal, 2016];

(3) Generating a large number of texts with artificially created OCR errors by inserting generic OCR errors and earlier detected period-specific OCR errors according to their determined frequencies into correct texts [Grouin, 2017] ;

(4) Training a neural network on the generated dataset.

Figure 1 illustrates the training dataset generation process for OCR post-correction based on a minimal amount of manually fixed texts.

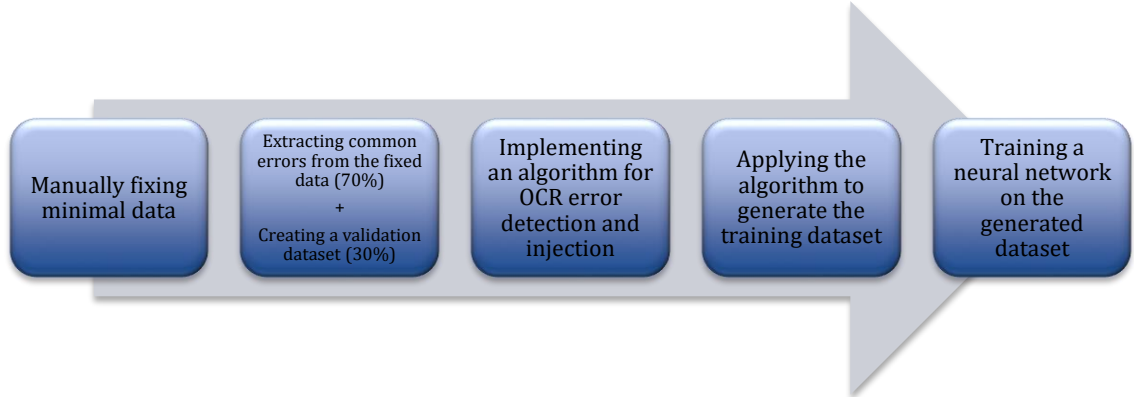

Figure 1: Corpus generation for neural network training with minimal manually fixed data

### 3.2 Hyperparameter Optimization

As NMT seems to be the closest task to the OCR error correction, we used the state-of-the-art NMT DNN's structure to build an effective DNN for the OCR error correction. Hence, the proposed network was designed as a sequence-to-sequence network [Sutskever, Vinyals, & Le, 2014] that is based on a recurrent neural network (RNN) with encoder-decoder architecture [Cho et al., 2014]. This model is suitable for matching a sentence with OCR errors (an input) to the same sentence without errors (an output). There are two main layer architectures in RNN sequence to sequence networks: Long Short Term Memory [Hochreiter & Schmidhuber, 1997] and Gated Recurrent Units [Cho et al., 2014]. Both LSTM and GRU are built from units. The number of units is a hyperparameter determined by the number of features that the layer can model ("learn"). Each layer has many more hyperparameters, but overall, there are fewer hyperparameters in GRU than in LSTM, which makes GRU more computationally efficient. However, LSTM can be more accurate for long sequences. It has been shown that there is no single optimal layer for all scenarios, and the optimal layer depends on the task, training dataset, and computational resources (CPUs and GPUs) available for training [Chung, Gulcehre, Cho, & Bengio, 2014].

Therefore, due to the complexity and the large number of hyperparameters, there are tens of thousands of possible network structures to train and validate for tuning the parameters, which makes it impractical to find the optimal network model for the specified task. To expedite the optimization process, we used an AutoML method - a greedy algorithm that sorts the hyperparameters in decreasing order of their computational complexity and optimizes only one of them at a time, starting with the most complex parameter to compute. This is a variation of the submodular optimization algorithm introduced in [Jin et al., 2016]; the process attempts to minimize the time required to find the optimal parameters. Once an optimal value for a hyperparameter was found in a training phase, it remains fixed for all the following training phases. This method minimizes the computational effort and time of the optimization process.

## 4  EXPERIMENTAL DESIGN

### 4.1  Dataset Generation

To evaluate the proposed approach for network optimization, a series of experiments have been conducted. The main goal of the experiments was to apply the developed methodology for building an optimized neural network for OCR post-correction in Hebrew historical newspapers. To this end, a test dataset (denoted as JP_CE) was created from 150 OCRed historical Hebrew newspaper articles from the JPress collection - the most extensive OCRed historical Hebrew newspapers collection dated 1800-2015. These articles were randomly selected from six newspapers in JPress with fine segmentation. As expected, the articles included OCR errors that were manually fixed by 75 volunteers, as described in the previous work [Suissa et al., 2020]. These articles were validated and double-checked by an expert to ensure correctness.

30% of the articles in this dataset were exploited for evaluating the network quality, while 70% were used to learn common OCR errors for the JPress corpus. In particular, to learn the most common character confusion pairs, 70% of the original JP_CE articles and their fixed golden standard versions were compared using the Needleman–Wunsch alignment algorithm [Needleman & Wunsch, 1970]. We note that these 70% of the dataset, comprised of 105 (of 150) articles with 2725 sentences, was the only manually corrected data required to train the constructed network. The most common OCR character confusion pairs are displayed in Table 1.

Table 1: Common OCR errors in Hebrew historical newspapers in JPress

| Character | Fix | Distribution |
|---|---|---|
| ח | ה | 8.17% |
| ד | ר | 5.01% |
| ג | נ | 4.19% |
| ב | כ | 3.44% |
| ׳ | , | 3.39% |
| ם | ס | 3.18% |
| ח | ת | 2.65% |
| ו | ׳ | 2.65% |

Next, four training datasets were composed as follows. To explore the influence of the genre and historical period of the training dataset on the correction accuracy of the resulted network, two datasets were based on texts from the Ben-Yehuda Project (the Hebrew equivalent of the Gutenberg project comprised of secular Hebrew literature mostly from the last two centuries and the Middle ages), and two other datasets consisted of the Hebrew Bible text. Thus, each pair of datasets belongs to a different time period and genre, while Ben-Yehuda's period (partially) overlaps with that of the JPress corpus. Both Ben-Yehuda and Bible texts were typed manually and are thus considered correct. To train the network, an OCRed version is needed for each of the datasets. These were created by inserting artificially generated OCR errors in each of the datasets. To compare the impact of the inserted error types on the obtained network, one of the Ben-Yehuda datasets (denoted as BYP) and one of the Bible datasets (denoted as BIBLE) was injected with artificially generated generic errors [Reynaert, 2008, Grouin, 2017; Jatowt et al., 2019] with a 20% noise ratio. In this case, characters in each line of the text (selected according to their relative frequency in Hebrew Wikipedia) were removed from words, inserted into words, or swapped with their consecutive characters as in [Grouin, 2017; Jatowt et al., 2019]. The second Ben-Yehuda dataset (denoted as BYP_HEB) and second Bible dataset (denoted as BIBLE_HEB) were injected with period-specific OCR errors and their relative frequencies learned from JPress as described above, in addition to the generic error insertion, with a 20% noise ratio. The pseudo-code of the error generation algorithm is displayed in Algorithm 1 below. As can be observed from Algorithm 1, first, the algorithm generates some generic error types. Next (lines 13-17 in Algorithm 1), we introduce a period-specific phase, where the most common JPress-specific OCR errors are added according to the relative frequency of their occurrence in the corpus. Table 2 summarizes the five datasets that were generated and utilized in this study and their main characteristics.

Table 2: The constructed datasets of the study

| Dataset | Input Corpus | Target Golden Standard Corpus | Generation method | Size (sentences) |
|---|---|---|---|---|
| JP_CE | JPress – OCRed historical newspapers | Fixed JPress articles | Manually fixed | 2,725 |
| BYP | The Ben-Yehuda Project with generic OCR errors | The Ben-Yehuda Project – books | Automatically inserted errors | 2,638,056 |
| BYP_HEB | The Ben-Yehuda Project Hebrew JPress specific OCR errors | The Ben-Yehuda Project - books | Automatically inserted errors | 2,638,056 |
| BIBLE | The Bible with generic OCR errors | The Hebrew Bible from sefaria.org.il | Automatically inserted errors | 527,156 |
| BIBLE_HEB | The Bible with Hebrew JPress specific OCR errors | The Hebrew Bible from sefaria.org.il | Automatically inserted errors | 527,156 |

ALGORITHM 1: Generation of generic and period-specific Hebrew OCR Errors

Input: Gold standard (GSt), Noise ratio (NR), Period-specific Hebrew errors (HebrewErrors), 0 < NR < 1
Output: OCRed artificial text (OCRt)
for each line in GSt do
    if random(0,1) < NR then
        Remove one character (selected according to Wikipedia-based frequency) from the line
    end if
    if random(0,1) < NR then
        Insert one character (selected according to Wikipedia-based frequency) into one random
        position in the line
    end if
    if random(0,1) < NR then
        for i in random(1,2) do
            Swap two consecutive characters at one random position in the line
        end for
    end if
    for each char in HebrewErrors do
        if GSt contains char and random(0,1) < NR*EP then (EP = Error Probability)

```
                                Replace char at one random position from all fixed char in GSt
                    end if
                end for each
                OCRt <= OCRt + line
end for each
return OCRt
```

## 4.2 Building Network Models

To build the network models for the study, the state-of-the-art network model from previous research was adopted as a baseline [Ghosh & Kristensson, 2017]. Table 3 illustrates the structure of the baseline neural network and its hyperparameters. We note that this model was found optimal for a different language and historical period (modern English) than those investigated in the current research. Then, several hyperparameters of the neural network were tuned by the greedy algorithm, in order to optimize the accuracy of the OCR correction and minimize the loss in the following order: 1) the number of layers, 2) the type of the layers, 3) the dropout level, 4) the number of units, 5) the batch size, and 6) the epoch size. The number of layers and the type of the layer increases the number of calculations that the network performs (as each layer adds neurons), while the dropout reduces the number of calculations (i.e., GPU needed). The number of users and the batch size increase the memory needed (i.e., GPU RAM), and the epoch size only influences the duration of the training process.

During the optimization process, 45 different network structures have been tested, and an optimal structure was found using the BYP_HEB dataset, as specified in Table 3. All networks were trained using Adam optimizer [Kingma & Ba, 2014] to optimize the learning rate during training, and used categorical cross-entropy loss function. The network's input was a single sentence with OCR errors. Using the greedy approach, the whole process took four months, while computing all the possible network structures with different values of the above hyperparameters would take over 50 years (using the same computational power). This result emphasizes the importance of the optimization of neural networks for specific time and language. Finally, the BYP_HEB training corpus described above was utilized to build two network models: 1) a model with the baseline structure and 2) a model with the optimized structure (denoted as BYP_HEB_DNN).

Table 3: The structure of the baseline network vs. the optimized network built for the BYP dataset

| Hyperparameter | Baseline value | Optimized Value |
| --- | --- | --- |
| Main Layer | GRU | LSTM |
| Depth | 2 (one in the encoder side and one in the decoder side) | 4 (two in the encoder side and two in the decoder side) |
| Dropout | 0.2 | 0.2 |
| Units | 500 | 500 |
| Epoch size | 20,000 | 250,000 |
| Bidirectional | No | Yes |
| Batch size | 100 | 256 |

## 5 EVALUATION

The suggested methodology and optimized model (BYP_HEB_DNN) were first applied on Hebrew BYP_HEB dataset (with period-specific OCR errors) described in section 4.1 and evaluated compared to: (1) OpenNMT-CLAM - another NMT methodology (i.e., one of the best models developed by the CLAM team from ICDAR 2019 [Rigaud, et al., 2019]), (2) the suggested model trained on BYP with modern language errors (i.e., extracted from texts from a different period), (3) the suggested model trained on ancient texts (BIBLE_HEB, the Bible) from a different period, and (4) industry-leading spell checkers (Microsoft Word and Google Docs, that are common for Hebrew).

In addition, Polish was chosen to evaluate the performance of the suggested optimized model on other morphologically-rich languages due to its relative complexity [Kulsum-Binder & Bahloul, 2016] and dataset availability. The best proposed model found for Hebrew (BYP_HEB_DNN) was trained on the ICDAR 2019 Polish training dataset [Rigaud, et al., 2019] with the same architecture and optimized hyper-parameter values that were used for Hebrew and with period-specific errors. This model is denoted throughout this paper as BYP_POL_DNN. To create a period-specific Polish dataset (denoted as POL_Period) for training this model, period-specific errors were extracted from the ICDAR 2019 Polish training dataset using Algorithm 1. The ICDAR 2019 Polish dataset was built using texts from the National Library of Poland (mostly historical newspapers). To assess the performance of the proposed approach based on period-specific errors vs. training on period-agnostic errors, the suggested model was also trained using the same historical Polish corpus but with errors extracted from modern texts (Polish Wikipedia); this dataset is denoted as POL_Generic. Then, OpenNMT-CLAM was applied on both POL_Period and POL_Generic, as a comparative evaluation of BYP_POL_DNN with other methods.

### 5.1 Evaluation Metrics

The standard accuracy evaluation metric of neural networks is defined as a percentage of the correct samples (perfectly correct OCRed sentences in our case) that the network produces out of all the samples processed by the network. This type of accuracy evaluation does not consider partially correct samples when the network was able to fix only some of the OCR errors present in the sentence, but not all of them. Hence, to assess the performance of the networks in a more comprehensive manner, two new evaluation measures were defined: 1) the added accuracy (Acc_Char) that estimates the increase in the character-based accuracy of the text corrected by the network, and 2) the overall word error rate (WER) and character error rate (CER) of the text corrected by the network.

The character-based metric is computed as a percentage of the errors fixed by the network out of the total number of OCR errors in the input text. The number of network's corrections is calculated as a difference between the Levenshtein's minimal edit distance [Levenshtein, 1966], denoted as *lev*, of the input OCRed text from the (correct) golden standard version of the text, GS, and the minimal edit distance of the fixed text, Fixed, (after the network's corrections) from GS. The initial number of errors in the OCRed text is computed as the minimal edit distance between the OCRed text and the golden standard text. If a network has inserted more errors than it has fixed, the accuracy metric value is set to 0. More formally, the Acc-Char metric is defined as follows:

$$Acc\_Char = \begin{cases} \frac{lev_{GS,OCRed} - lev_{GS,Fixed}}{lev_{GS,OCRed}} * 100, & lev_{GS,OCRed} \geq lev_{GS,Fixed} \\ 0, & otherwise. \end{cases}$$

As a preparatory step for estimating the accuracy of the given text at the word-level, the Needleman-Wunsch alignment algorithm [Needleman & Wunsch, 1970] was applied to compare the evaluated text with its golden standard version. Then, the alignment's output was processed to split the text into words using a standard set of delimiters. The word-based accuracy of the text compared to its golden standard version is assessed with the standard accuracy measure [Ali & Renals, 2018]:

$$\text{WER} = \frac{I_w + S_w + D_w}{N_w}$$

where $N_w$ is the total number of words in the evaluated text, $S_w$ is the number of words in the evaluated text that are substituted with other words in the golden standard version of the text, $I_w$ is the number of words in the evaluated text that are absent from the golden standard text, and $D_w$ is the number of words that occur in the golden standard text but are absent from the evaluated text. We also measure the standard CER measure, which is calculated in the same manner as the WER measure but at a character level. The word-based metric complements the character-based accuracy metric, as it recognizes a correction as a good one only if all the errors in a word were fixed. This measure is crucial from the user perspective since users comprehend and search texts by whole words.

Figure 2 summarizes the proposed approach for optimizing the neural network's training process for OCR error correction and its evaluation.

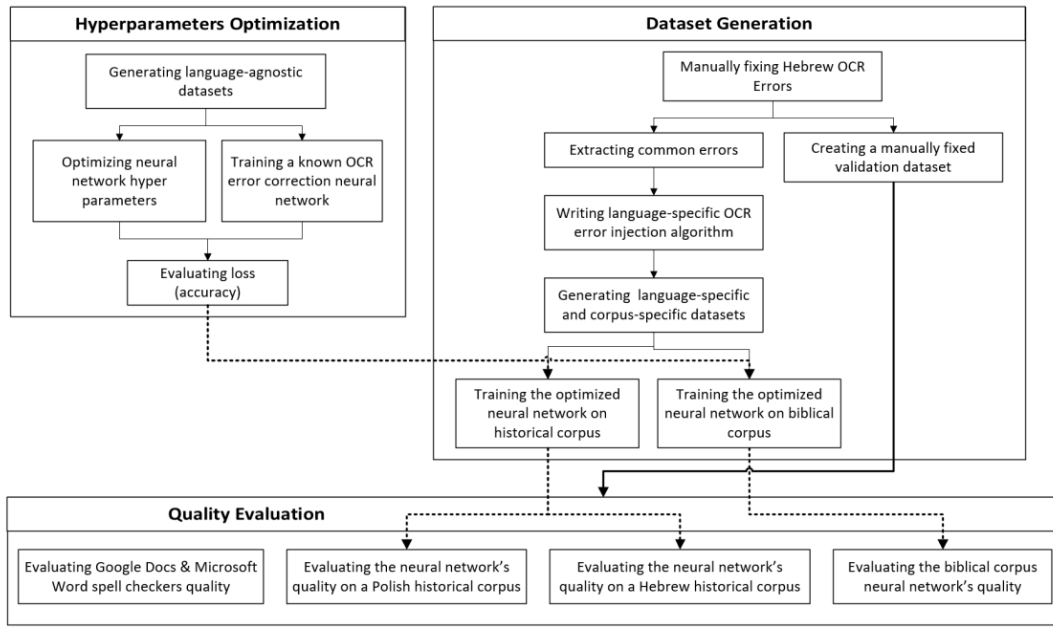

Figure 2: The summary of the proposed methodology.

## 6 RESULTS

### 6.1 Quantitative evaluation of the networks' performance and its influence factors

First, we compared the standard validation accuracy of the network with a baseline (GRU-based) structure to that with the optimized (LSTM-based) structure trained on the same dataset (BYP). This analysis aimed to assess the networks' ability to correct artificially generated OCR errors in new articles from the corpus they were trained on. Hence, the networks were trained on 80% of the sentences in the corpus (training set) and validated on the rest 20% sentences (validation set) from the same corpus.

We found that the best performing network was the optimized model trained using the greedy approach. In particular, the optimized network outperformed the baseline network by 3% (85% vs. 88% validation accuracy), as shown in Figure 3.

The individual hyperparameter's impact on the network accuracy was as follows:
- The dropout level (without vs. with 0.2-0.5) improved the network's accuracy by 2.29%.
- The number of layers (GRU with two layers vs. GRU with four layers) improved the network's accuracy by 2.09%.
- The batch size (32-512) improved the network's accuracy by 1.88%.
- The bidirectionality (without vs. with) improved the network's accuracy by 0.57%.
- The epoch size (5000-250000) improved the network's accuracy by 0.27%.
- The type of the layers (Bidirectional-GRU vs. Bidirectional-LSTM) improved the network's accuracy by 0.13%.
- The number of units (200-1000) did not improve the network's accuracy significantly.

Note that some of the improvements are overlapping; therefore, the total improvement was 3%, as shown in Figure 3. One of the open questions intensively discussed in the literature is the effect of the network architecture on the performance for various tasks [Yin, Kann, Yu & Schütze, 2017]. In this regard, it is interesting to note that the type of the layers (i.e., the network architecture) had only a minor effect on the performance of the OCR post-correction network in historical Hebrew texts. In addition, the results demonstrate a positive impact of period-specific OCR errors on the networks' correction accuracy. Accordingly, Figure 4 shows that the optimized network that was trained and validated on the BYP_HEB dataset (with JPress-specific OCR errors) reached a higher accuracy (94%) than the same network trained and validated on BYP (with generic OCR errors) (88%). We also found that after 256 sentences in a signal batch, the accuracy enters a plateau, and there is no more improvement. This means that for high-resource scenarios, there is a limited benefit using this optimization.

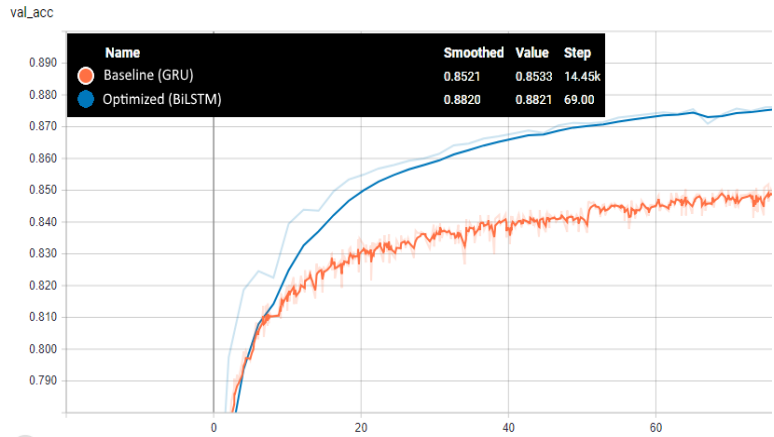

Figure 3: Baseline and optimized networks' validation accuracy for BYP_HEB. The orange line shows the baseline network validation accuracy, while the blue line displays the optimized network validation accuracy, x-axis: number of epochs, y-axis: validation accuracy from 0 to 1.

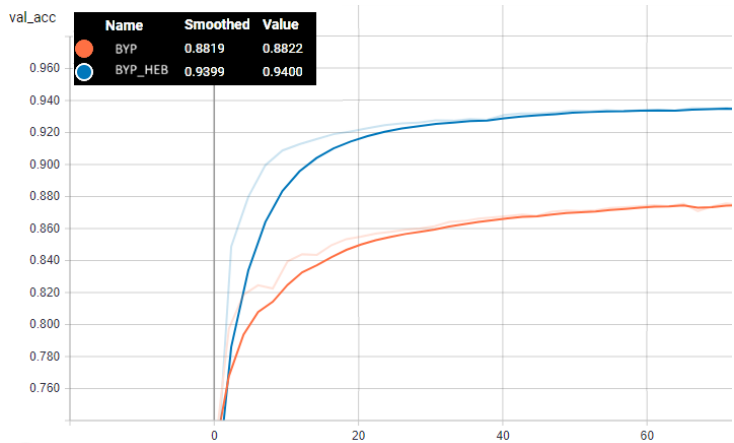

Figure 4: The optimized networks' validation accuracy for BYP and BYP_HEB. The orange line shows the network validation accuracy on BYP, while the blue line displays the network validation accuracy on BYP_HEB, x-axis: number of epochs, y-axis: validation accuracy from 0 to 1.

Next, we evaluated the network's ability to correct real OCR errors in the study's test dataset, JP_CE (selected historical newspapers from JPress). The additional challenge, in this case, is that the networks were tested on a dataset from a different genre and/or historical period than the corpora they had been trained on. At this stage, we examined only the two best networks (optimized models trained on BYP_HEB and BIBLE_HEB) and compared their results to those of the state-of-the-art NMT model: OpenNMT [Klein, Kim, Deng, Senellart & Rush, 2017] and the state-of-the-art spellchecking networks available on the market: the Google Docs and Microsoft Word 2019. OpenNMT is often used as a benchmark for OCR post-correction [Hakala et al., 2019; Rigaud et al., 2019]. We followed the CLAM team method for training the OpenNMT-CLAM model [Rigaud, et al., 2019], using [Grouin, 2017] error injection method (that is similar to Algorithm 3.1 but without the period-specific phase). State-of-the-art spellchecking networks often used by end-users (due to a lack of OCR post-correction tools in Hebrew) are based on (modern) language models in Hebrew and considered to be very good in text correction. Note that the initial word-based accuracy (calculated as: $(1 - WER) * 100$) of the original test dataset (JP_CE) was 48.984% (i.e., about 51% of the words were incorrect before applying the networks), and the character-based accuracy (calculated as: $(1 - CER) * 100$) was 91.439% (i.e., about 8.5% of the characters were incorrect).

The evaluation results are presented in Table 4. The obtained results indicate the dependency of the network's effectiveness on the period of the training dataset. When the network learns from the corpus from a different genre but written in a similar period, it achieves positive and much better results (around 4.5% character-based and 5.5% word-based accuracy increase), than networks trained on texts from substantially more distant periods (with none or negative change in the accuracy). In absolute terms, the network corrected about 9% (5% out of 51%) of the errors in the test dataset. To test the statistical significance of the results, a McNemar statistical test, suitable for small datasets, was calculated [Dietterich, 1998]. The test was applied to compare BYP_HEB_DNN to each of the other network models, and all the differences were found statistically significant.

Table 4: Networks' performance evaluated on JP_CE

| Network/model | Character-based Accuracy Increase (in %) | 1-WER (in %) | 1-CER (in %) | P-Value | χ2 distribution |
|---|---|---|---|---|---|
| BYP_HEB_DNN | 5.41 | 53.47 | 92.12 | | |
| Google Docs spell checker | ~ 0.00 | 41.58 | 90.08 | 0.001 | 68.445 |
| Microsoft Word spell checker | ~ 0.00 | 41.53 | 90.18 | 0.001 | 68.445 |
| OpenNMT-CLAM | ~ 0.00 | 34.36 | 86.85 | 0.000 | 64.557 |
| BIBLE_HEB_DNN | ~ 0.00 | ~ 0.00 | 72.89 | 0.000 | 51.040 |

When comparing the contribution (added accuracy) of the neural networks, it seems that the neural network that learns from the Bible did not improve the OCRed text at all. This is similar to a Hebrew-speaking person transported in time from the Biblical era to the 19th century and trying to read a morning Hebrew newspaper.

For similar reasons, the modern language NMT and the spellcheckers of Google Docs and Microsoft Word 2019 failed to fix historical Hebrew OCRed texts. Their added accuracy score (Acc_Char) was actually below 0 since they created more errors than they fixed. The spellcheckers are based on deep neural networks trained on the massive amount of Hebrew texts available online. As a result, the spellcheckers have a strong bias towards the modern Hebrew language that differs from the historical Hebrew of the JPress newspaper collection.

In this study, we experimented with a 10%-50% noise ratio to test the effect of the noise on the network's accuracy. As shown in Figure 5, the amount of noise affects the network performance (1-CER in %). At the noise level of 30%, the network's accuracy reduces, but enters a plateau up to 50%.

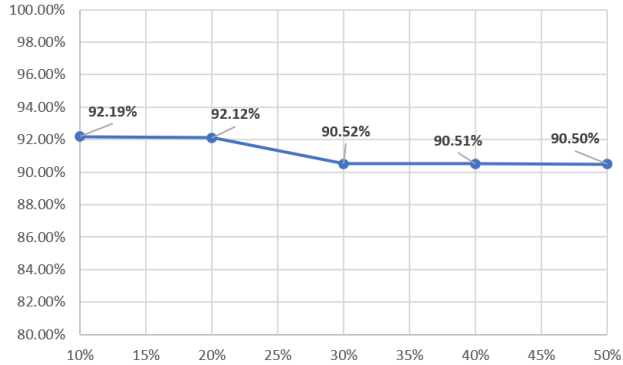

Figure 5: The noise ratio impact on the BYP_HEB_DNN accuracy (1-CER in %).

*6.1.1 Comparative evaluation of the proposed model on Polish datasets*

To compare the suggested optimized model's contribution to other morphologically-rich languages, its performance was evaluated on two Polish datasets (described in section 5). As can be observed from Table 5, the models that were trained on POL_Period (a dataset with period-specific errors) achieved better accuracy. This finding shows the value of period-specific errors in other morphologically-rich languages. Moreover, the fact that the BYP_POL_DNN achieved higher accuracy than the OpenNMT-CLAM shows that the hyper-parameters found for Hebrew improve accuracy on other morphologically-rich languages. The ICDAR 2019 Polish evaluation dataset's WER and CER were 26.5% and 61.1%, respectively. Furthermore, the character-based accuracy measurement using the suggested model was higher on Polish (44.8%) than on Hebrew (5.41%); this could be due to the low accuracy of the Polish validation dataset (61.1%) compared to the Hebrew validation dataset (91.4%) (i.e., there was more room for improvement in the Polish dataset). Another reason could be the complexity of the Hebrew language (i.e., Hebrew is more difficult to fix).

Table 5: Networks' performance evaluated on ICDAR 2019 Polish evaluation dataset

| Network/model | Training dataset | Character-based Accuracy Increase (in %) | 1-WER (in %) | 1-CER (in %) |
|---|---|---|---|---|
| BYP_POL_DNN | POL_Period | 44.88 | 51.98 | 78.56 |
| OpenNMT-CLAM | POL_Period | 36.09 | 50.22 | 75.14 |
| BYP_POL_DNN | POL_Generic | ~ 0.00 | 26.46 | 61.12 |
| OpenNMT-CLAM | POL_Generic | ~ 0.00 | 26.41 | 61.11 |

## 6.2 Qualitative analysis of the networks' performance

To qualitatively assess the network's performance, we manually analyzed 20% of randomly selected corrections. We found that the best optimized neural network of the study (trained on BYP_HEB) was able to fix the following error types:

- A single wrong character in a word;
- Multiple wrong characters in a word;
- Swapped characters in a word;
- Redundant white spaces;

- Redundant characters;
- Missing white spaces.

Figure 6 shows the distribution of the corrected OCR error types.

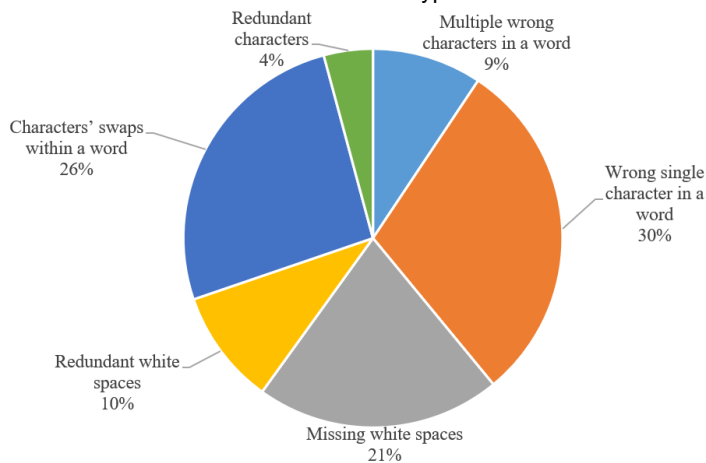

Figure 6: The distribution of the common OCR error types in Hebrew historical newspapers in JPress.

Table 6 below displays examples of the above error types and the corresponding network's corrections.

Table 6: Examples of error types detected and corrected by the network

| Error | OCRed Text | BYP_HEB_DNN Correction |
| --- | --- | --- |
| Character replacement error | הילד התחיל בוכה מתוg פחד | הילד התחיל בוכה מתוך פחד |
| Character swap error | היא אומרת לו, היסו תקף | היא אומרת לו, היום תקף |
| Removing and adding a white space | לפעמיםה מה מתנגשים איש באחיו ודוחפים | לפעמים המה מתנגשים איש באחיו ודוחפים |
| Removing a redundant character | איה tהמלאכים באו | איה המלאכים באו |

Remarkably, the network was also able to correct errors in GS (golden standard text). Although GS was created manually and double-checked by an expert, it contained some errors (less than 1%, mainly related to spacing) that were fixed by the network, as illustrated in Figure 7. In the presented example, the network added a missing space between the colon and the next word, which did not appear in the OCRed or GS texts. This indicates that the network learned a comprehensive language model of the given corpus, rather than just a simple mapping of the common errors.

> **OCR**: היא פשוטה:הפועלים מתיאשים מפני שאינם רואים
> **GS**: היא פשוטה:הפועלים מתיאשים מפני שאינם רואים
> **BYP_HEB_DNN**: היא פשוטה: הפועלים מתיאשים מפני שאינם רואים

Figure 7: An example of the GS's error correction by the BYP_HEB DNN.

We also analyzed the optimized network (BYP_HEB_DNN) results compared to those of the industry-leading spellcheckers and the OpenNMT-CLAM. As can be observed in Figure 8, the developed, optimized neural network fixed both errors in the given example, while the leading spellcheckers missed the first error but fixed the second one, and the NMT missed both fixes. The difference between the two errors is that the first one is a real word (namely the card "ace" in Hebrew), which does not fit in the specified context, while the second one is not a real word and thus cannot be found in the Hebrew dictionary. Furthermore, while MS Word's correction was right (replacement of the last letter of the word – **י** with a similarly looking **ד**), Google Docs' correction inserted a new first letter **כ** and created a different real word **כיסוי** (**coverage** in Hebrew). Although this correction fits in the context and has the same edit distance to the original word as the right correction, it is clearly the wrong word.

> **OCR**: א**ס** יש **יסוי** כלשהו להצהרותיהם והבטחותיהם
> **GS**: א**ם** יש יסו**ד** כלשהו להצהרותיהם והבטחותיהם
> **OpenNMT-CLAM**: א**ס** יש **יסוי** כלשהו להצהרותיהם והבטחותיהם
> **MS Word**: א**ס** יש יסו**ד** כלשהו להצהרותיהם והבטחותיהם
> **Google Docs**: א**ס** יש **כ**יסוי כלשהו להצהרותיהם והבטחותיהם
> **BYP_HEB_DNN**: א**ם** יש יסו**ד** כלשהו להצהרותיהם והבטחותיהם

Figure 8: An example of wrong OCR fix by OpenNMT-CLAM, leading spellcheckers, and the correct fix by BYP_HEB_DNN.

As can be observed from the example in Figure 9, the neural network also outperforms the spellcheckers in the identification and preservation of named entities (e.g., names of places, people, and events). Both spellcheckers tried to fix the correct word **מקנטרבארי** (from Canterbury in Hebrew), since its modern spelling differs from the historical spelling in the OCRed text. The NMT was also able to preserve the entities. This is another essential advantage of the network trained on the text from a similar historical period to the target OCRed text, which can critically impact the searchability of the corrected OCRed text.

> **OCR**: הארכיבישוף **מקנטרבארי** והארכיבישוף **מיורק**
> **GS**: הארכיבישוף **מקנטרבארי** והארכיבישוף **מיורק**
> **OpenNMT-CLAM**: הארכיבישוף **מקנטרבארי** והארכיבישוף **מיורק**
> **MS Word**: הארכיבישוף מקנט**בריה** והארכיבישוף **מיורק**
> **Google Docs**: הארכיבישוף מקנטרב**רי ה**ארכיבישוף **מיורק**
> **BYP_HEB DNN**: הארכיבישוף **מקנטרבארי** והארכיבישוף **מיורק**

Figure 9: Preserving the names of entities.

In addition, the optimized network mostly detected and fixed punctuation errors, while both spellcheckers and the NMT usually ignored the wrong punctuation, as shown in the example in Figure 10.

> **OCR**: להצטרף ליחידות וואלשיו**ת .** החריצות**'**  העבודה והרוח הקיבוצית
> **GS**: להצטרף ליחידות וואלשיו**ת,** החריצות, העבודה והרוח הקיבוצית
> **OpenNMT-CLAM**: להצטרף ליחידות וואלשיו**ת .** החריצו**ת .** העבודה והרוח הקיבוצית
> **MS Word**: להצטרף ליחידות ו**ו**לשיו**ת .** החריצות**'** העבודה והרוח הקיבוצית
> **Google Docs**: להצטרף ליחידות ו**ו**לשיו**ת .** החריצות**'** העבודה והרוח הקיבוצית
> **BYP_HEB DNN**: להצטרף ליחידות וואלשיו**ת,** החריצות, העבודה והרוח הקיבוצית

Figure 10: Punctuation correction example.

Moreover, the spellcheckers added new errors that did not exist in the original OCRed text, especially in cases of grammatical conjugations. In the example in Figure 11, MS Word wrongly replaced the last letter in the word **ללמודי** (to studies), while Google Docs completely removed the last letter of this word.

> **OCR**: תקבענה בו כמה שעות ללמודי הד**ת .**
> **GS**: תקבענה בו כמה שעות ללמודי הד**ת.**
> **OpenNMT-CLAM**: תקבענה בו כמה שעות ללמודי הד**ת.**
> **MS Word**: תקבענה בו כמה שעות ללמוד**ה** הד**ת .**
> **Google Docs**: תקבענה בו כמה שעות ללמו**ד** הד**ת .**
> **BYP_HEB DNN**: תקבענה בו כמה שעות ללמודי הד**ת.**

Figure 11: New spelling mistakes created by the spellcheckers.

In summary, from an in-depth manual examination of 20% of randomly selected corrections, it seems that the spellcheckers' corrections are strongly lexicon-focused as they successfully detected and fixed non-real words that do not exist in the lexicon of the spellcheckers. For the same reason, they tried to fix valid entities' names that were not found in the modern Hebrew lexicon of the spellcheckers due to a different spelling in the historical text. However, they tend to ignore the context of the corrected words. Thus, they failed to identify real‐word errors that do appear in their lexicon, but do not make sense in the given context. In addition, non-real words have always been corrected by the spellcheckers, but the chosen corrections did not always semantically match the context. Using a modern NMT with an optimal training approach in other languages (CLAM team's model in ICDAR 2019) successfully preserved contextual words, such as named entities, but failed to fix real words and some non-real words. The NMT seems to overfit on the small dataset and was unable to capture the rich morphology of the Hebrew language. We anticipate that optimizing OpenNMT (regardless of the CLAM method) as was performed for the BYP_HEB_DNN will probably achieve better results, since OpenNMT is based on the same architecture as BYP_HEB_DNN. Table 7 summarizes the findings of the error analysis based on the manual examination of the corrected test dataset compared to the original test dataset. Overall, the findings presented in the table demonstrate that the proposed design of a neural network trained on the text from a close historical period (but from a different genre) improves the OCR correction accuracy compared to other DNN models.

Table 7: Error types resolved (% of errors corrected) by the optimized neural network vs. industry spellcheckers and NMT.

| Error | | BYP_HEB_DNN | Industry spellcheckers | OpenNMT-CLAM |
|---|---|---|---|---|
| Single character / Multiple characters | Real word | ✓ (~95%) | ✗ (~5%) | ✗ (~25%) |
| | Non-real word | ✓ (~95%) | ✓ (~95%) | ✓ (~55%) |
| Missing spacing | | ✓ (~100%) | ✗ (~20%) | ✗ (~35%) |
| Redundant spacing | | ✓ (~100%) | ✓ (~55%) | ✗ (~20%) |
| Characters swaps | Real word | ✓ (~90%) | ✗ (~0%) | ✗ (~25%) |
| | Non-real word | ✓ (~95%) | ✓ (~100%) | ✓ (~95%) |
| Preserving the names of entities | | ✓ (95%) | ✗ (15%) | ✓ (95%) |

## 7 CONCLUSIONS

This study proposed and implemented a multi-phase optimization method for training an effective DNN for automatic OCR error correction in historical corpora. The developed method tackles several challenges of this complex task, such as the lack of training data and optimal DNN for historical texts in morphologically-rich languages, a large number of hyperparameters and network structures to examine and tune, the need for large manually corrected training datasets, and time-consuming optimization and training process.

The method was applied for training DNN networks on four historical corpora with different linguistic characteristics and artificially generated OCR error types. The performance of the obtained networks was then comparatively evaluated by OCR error correction of historical Hebrew newspapers in JPress. The optimized neural network, based on LSTM and trained on the Ben-Yehuda corpus with period-specific artificial OCR errors, yielded 94% accuracy compared to only 85% accuracy produced by the baseline GRU-based network on the Ben-Yehuda dataset with generic OCR errors. Interestingly, when trained on the very ancient (such as the Bible) or modern texts (as in the case of industry spellcheckers and OpenNMT-CLAM), the neural network performed quite poorly and sometimes even introduced more errors than corrections. This can be explained by the significant change in the Hebrew language over time that emphasizes the importance of a period-specific approach. However, it seems that texts from different genres have enough in common to achieve good results when training the model on an available corpus and applying it to correct texts from another corpus from a different genre (as was the case for the BYP and the JP datasets). Moreover, the suggested model is lightweight and can be trained within a few hours, while industry-leading spellcheckers and OpenNMT-CLAM are complex and require significantly more resources and time. Remarkably, only 150 manually corrected articles were enough to learn the period-specific error types and substantially augment the training datasets. The optimized model was also found effective for other morphologically-rich languages, and substantially improved the performance on the historical Polish dataset.

In summary, this paper's contributions are (1) a period-specific approach for morphologically-rich languages and specifically, the detection and classification of common OCR error types at the character level for historical Hebrew, (2) an algorithm for generation of large artificial training corpora with automatically inserted common period-specific OCR errors in Hebrew, (3) crowdsourcing a low resource dataset generation and the minimal use of the manually corrected data in corpus generation, (5) the quantitative analysis and evaluation of OCR error post-correction for Hebrew, and (4) a critical impact of the historical period of the training text on the network's accuracy.

The proposed methodology can be applied to historical and cultural OCRed collections in Hebrew and other morphologically-rich languages. Researches can utilize the study's results to reduce the time, cost, and complexity when designing neural networks for OCR post-correction and, thus, to improve the OCRed document correction process for many digital humanities projects. Future work may expand the proposed method to literary and temporally diverse historical corpora in Hebrew and other morphologically-rich languages, such as Arabic, and using transfer learning from different corpora.